\documentclass[conference]{IEEEtran}
\IEEEoverridecommandlockouts

\usepackage{cite}
\usepackage{amsmath,amssymb,amsfonts}
\usepackage{algorithmic}
\usepackage{graphicx}
\usepackage{textcomp}
\usepackage{xcolor}
\usepackage{multirow}
\usepackage{booktabs}

\def\BibTeX{{\rm B\kern-.05em{\sc i\kern-.025em b}\kern-.08em
    T\kern-.1667em\lower.7ex\hbox{E}\kern-.125emX}}
\begin{document}

\title{MADiff: Text-Guided Fashion Image Editing with Mask Prediction and Attention-Enhanced Diffusion\\

\thanks{* Equal Contribution. $\dag$ Corresponding author. This work was supported by Research Fund for Advanced Ocean Institute of Southeast University, Nantong (General Program;Key Program; Major Program), Guangdong Basic and Applied Basic Research Foundation(2022A1515011435), ZhiShan Scholar Program of Southeast University and the Fundamental Research Funds for the Central Universities, the Natural Science Basic Research Program of Shaanxi (Program No.2024JC-YBMS-513), and Key Research and Development Program of Zhejiang Province under Grants 2024C01025. }
}

\author{\IEEEauthorblockN{Zechao Zhan$^{1}$*, Dehong Gao$^{3, 4}$*, Jinxia Zhang$^{1, 2\dag}$, Jiale Huang$^{1}$, Yang Hu$^{1}$, Xin Wang$^{5}$}
\IEEEauthorblockA{\textit{$^{1}$Key Laboratory of Measurement and Control of CSE, Ministry of Education, }\\\textit{School of Automation, Southeast University, Nanjing 210096, China} \\
\textit{$^{2}$Advanced Ocean Institute of Southeast University, Nantong 226010, China}\\
\textit{$^{3}$Northwestern Polytechnical University, School of Cybersecurity, Xi'an, China} \\
\textit{$^{4}$Binjiang Institute of Artificial Intelligence, ZJUT, Hangzhou, China}\\     
\textit{$^{5}$Alibaba Group, Hangzhou, China} }

}
\maketitle

\begin{abstract}
Text-guided image editing model has achieved great success in general domain. However, directly applying these models to the fashion domain may encounter two issues: (1) Inaccurate localization of editing region; (2) Weak editing magnitude. To address these issues, the MADiff model is proposed. Specifically, to more accurately identify editing region, the MaskNet is proposed, in which the foreground region, densepose and mask prompts from large language model are fed into a lightweight UNet to predict the mask for editing region. To strengthen the editing magnitude, the Attention-Enhanced Diffusion Model is proposed, where the noise map, attention map, and the mask from MaskNet are fed into the proposed Attention Processor to produce a refined noise map. By integrating the refined noise map into the diffusion model, the edited image can better align with the target prompt. Given the absence of benchmarks in fashion image editing, we constructed a dataset named Fashion-E, comprising 28390 image-text pairs in the training set, and 2639 image-text pairs for four types of fashion tasks in the evaluation set. Extensive experiments on Fashion-E demonstrate that our proposed method can accurately predict the mask of editing region and significantly enhance editing magnitude in fashion image editing compared to the state-of-the-art methods.
\end{abstract}

\begin{IEEEkeywords}
Text-guided Image Editing, Fashion Domain, Diffusion Model
\end{IEEEkeywords}

\section{Introduction}
Due to advances in text-to-image models \cite{dalle2, rombach2022high, saharia2022photorealistic} and diffusion models \cite{ho2020denoising} \cite{song2020denoising}, text-guided image editing has also experienced rapid development. Blended Diffusion \cite{avrahami2022blended} and Blended Latent Diffusion \cite{avrahami2023blended} combine noisy versions of the original image with the intermediate results of the diffusion model to perform text-guided editing. Based on these framework, DiffEdit \cite{couairon2023diffedit} automatically generates masks for the editing regions by comparing the differences between two generation pipelines. To identify an accurate latent space similar to GAN Inversion \cite{brock2018large} for effective image editing, DDIM Inversion \cite{preechakul2022diffusion} is proposed for diffusion models. Based on this inversion space, Imagic \cite{kawar2023imagic} and Prompt Tuning Inversion \cite{dong2023prompt} optimize text prompts and use interpolation for image editing. Another type of models, such as P2P \cite{hertz2023prompttoprompt}, FPE \cite{brooks2023instructpix2pix}, and PnP \cite{tumanyan2023plug}, control the editing direction and intensity through modifications of the attention map. Instruct-pix2pix \cite{brooks2023instructpix2pix}, based on P2P \cite{hertz2023prompttoprompt} and GPT-3 \cite{NEURIPS2020_1457c0d6}, proposes an inversion-free model for text-guided editing based on instruction.

\begin{figure}
    \centering
    \includegraphics[width=0.9\linewidth]{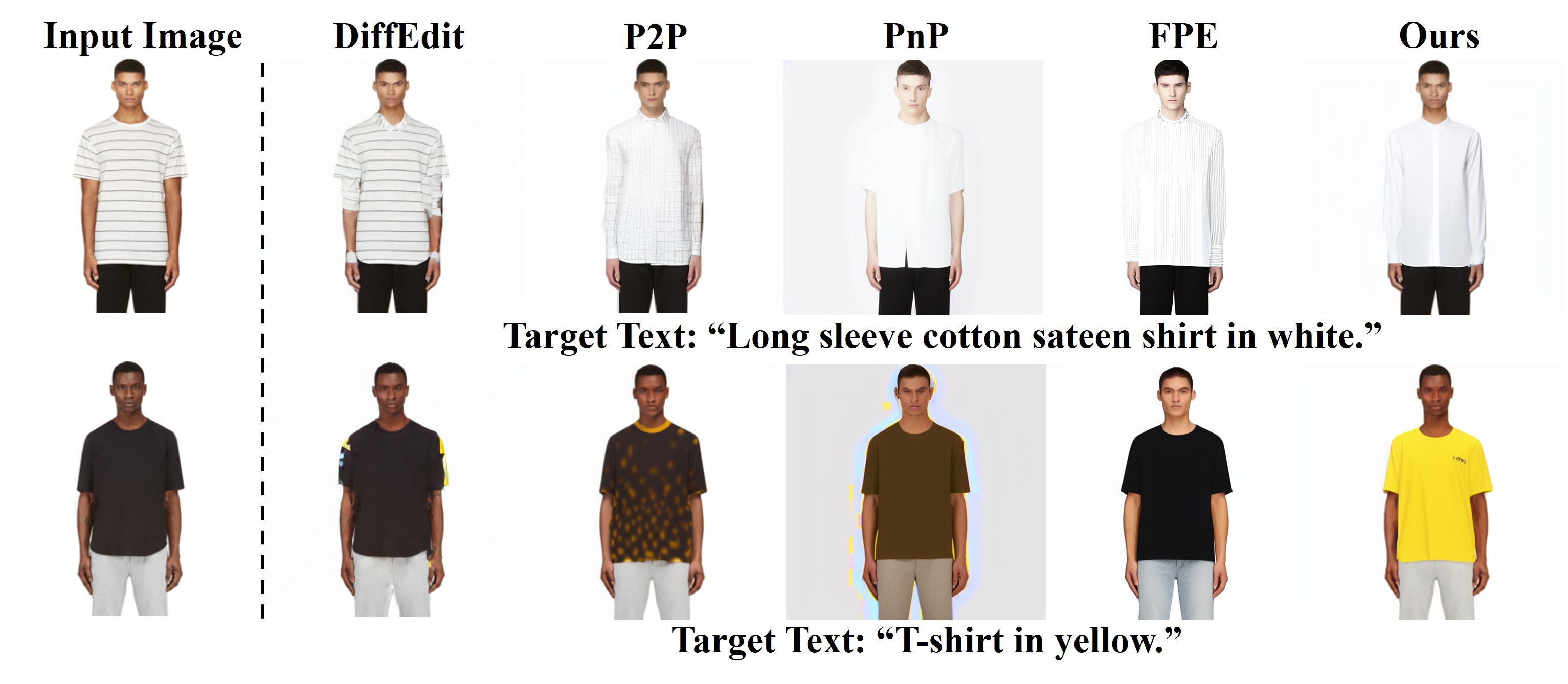}
    \vspace{-1.2em}
    \caption{General-domain text-guided image editing models have inaccurate localization of the editing region (e.g. wrongly handle the sleeve length) and weak editing magnitude (e.g. cannot change the color from black to yellow).  }
    \vspace{-1.5em}
    \label{fig:intro}
\end{figure}

Although existing text-guided editing models show promising results in general domain, they primarily focus on global edits, such as changing the object categories and styles, with generally weak editing magnitudes. In the fashion domain, model must possess sufficient editing magnitude to handle editing tasks with significant visual differences, as well as consider local editing tasks. Therefore, directly applying general-domain editing models to the fashion domain may encounter the following issues: (1) Inaccurate localization of editing regions; (2) Weak editing magnitudes. As shown in the first row of Fig. \ref{fig:intro}, DiffEdit \cite{couairon2023diffedit} and PnP \cite{tumanyan2023plug} fail to adequately handle edits related to sleeve length, while FPE \cite{liu2024towards} makes additional modifications to the face, reflecting inaccurate localization of editing regions. As illustrated in the second row of Fig. \ref{fig:intro}, the t-shirt edited  by general-domain methods still largely retain the original black color rather than the yellow color described in the target text, indicating weak editing magnitude.

\begin{figure*}
    \centering
    \includegraphics[width=0.7\linewidth]{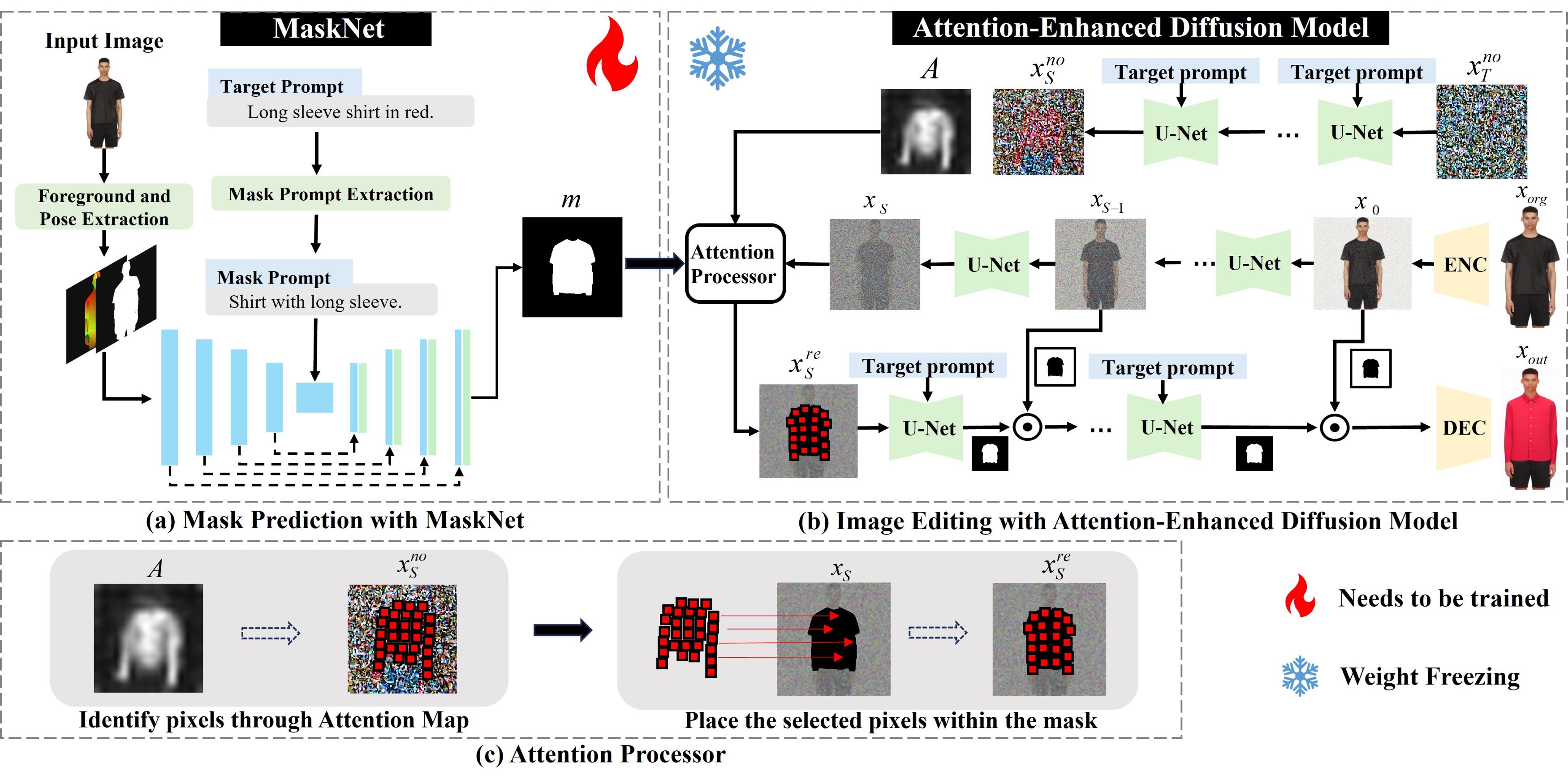}
    \vspace{-1.2em}
    \caption{Overview of our model. (a) Mask Prediction phase: The input image and target prompt are preprocessed to get the foreground region, densepose map and mask prompt, which is then input into MaskNet to predict the editing region. (b) Image Editing phase: DDIM inversion and sampling are conducted to get the attention and noise maps. Then all these maps and the mask from MaskNet are fed into the Attention Processor to create a refined noise map, which is finally used for blended editing to produce the edited image. (c) Attention Processor: The pixels with higher attention values are identified and then placed within the predicted editing region.}
    \vspace{-1.5em}
    \label{fig:method}
\end{figure*}

To address these issues, we propose the MADiff model, comprising of two main components: MaskNet and Attention-Enhanced Diffusion Model. Specifically, to address the issue of inaccurate localization of editing region, the MaskNet is proposed, in which the foreground region, the densepose map and the mask prompt from large language model are fed into a lightweight UNet to predict the mask for editing region. To strengthen the editing magnitude, the Attention-Enhanced Diffusion Model is proposed, where pixels with higher attention values are used to replace the original pixels within the editing region through the proposed Attention Processor. As shown in Fig. \ref{fig:intro}, compared to state-of-the-art editing models, our method shows accurate localization of editing region and sufficient editing magnitude.


Given the absence of benchmarks in fashion image editing, the Fashion-E dataset is constructed, which is composed of the training set and the evaluation set. The training set consists of 29380 aligned image-text pairs from Fashion-Gen dataset \cite{rostamzadeh2018fashion} and the corresponding cloth masks, which serve as the inputs and ground truth for the training of MaskNet respectively. The evaluation set is designed for four types of fashion editing tasks: color, detail, material, and comprehensive editing tasks. Within this set, 2639 images are annotated with task-specific target texts to comprehensively assess the editing models.

We conduct extensive experiments on Fashion-E and evaluate our model using a range of metrics. The experimental results indicate that our method not only predicts more accurate editing regions but also significantly enhances editing magnitude. Additionally, an ablation study is performed to demonstrate the effect of each component in our method. The contributions of this work can be summarized as follows:
\begin{itemize}
\item A MaskNet is proposed to accurately predict the mask of editing region, which significantly enhances the model's capability in handling local edits.
\item A novel Attention-Enhanced Diffusion Model is proposed to address the issue of weak editing magnitude in text-guided fashion image editing, enabling effective editing based on the target prompt.
\item  Given the absence of benchmarks in fashion image editing, a new dataset named Fashion-E is proposed, which supports the evaluation of different models on various fashion editing tasks. 
\item The experimental results on Fashion-E demonstrate that our model outperforms state-of-the-art models in both text alignment and preservation of original information.
\end{itemize}

\section{Method}
Our method consists of two main phases: mask prediction and image editing, as shown in Fig. \ref{fig:method}. In the mask prediction phase, input image and target prompt are first processed by Graphonomy \cite{gong2019graphonomy}, DensePose \cite{guler2018densepose} and LLAMA3-8b \cite{llama3modelcard} to get the foreground region, densepose map and mask prompt, which are then input into MaskNet to predict the mask of editing region. In the image editing phase, DDIM inversion from input image and DDIM sampling from random noise are first conducted to obtain the attention map, and noise maps. Then the attention map, noise maps and the mask from MaskNet, are input into the Attention Processor to produce a refined noise map. Finally, the refined noise map performs blended editing with the mask to obtain the edited image.

\vspace{-0.25em}
\subsection{Mask Prediction with MaskNet}
\vspace{-0.25em}
MaskNet is proposed to accurately predict the editing region based on the target prompt and the input image. As illustrated in Fig. \ref{fig:method}(a), MaskNet adopts a lightweight UNet as framework, in which the spatial attention layers are used in the middle block to incorporate text information. Graphonomy and DensePose are utilized to extract the foreground region and densepose map of the input image, which are then concatenated and applied as the input of the MaskNet. Additionally, considering that the editing region is only decided by the shape-related vocabulary in the target prompt, LLAMA3-8b is utilized to process the target prompt, yielding a mask prompt that only retains the words describing the shape of the fashion object. 
The MSE Loss between the predicted mask and the cloth mask, which is aligned with the text, is used to train MaskNet on the training set of Fashion-E.

The key distinction between MaskNet and other text-guided segmentation models \cite{shagidanov2024grounded, wang2024texfit, ren2024grounded} is that MaskNet does not simply segment objects from fashion images. As shown in Fig. \ref{fig:method}, with the help of the target prompt, MaskNet can accurately predict the editing region, such as the long sleeve region requested in the target text, rather than only the short sleeve t-shirt region. This enables our model to perform editing tasks that involve altering the shape of fashion objects, such as modifying sleeve length and collar shape. 


\begin{figure}
    \centering
    \includegraphics[width=0.75\linewidth]{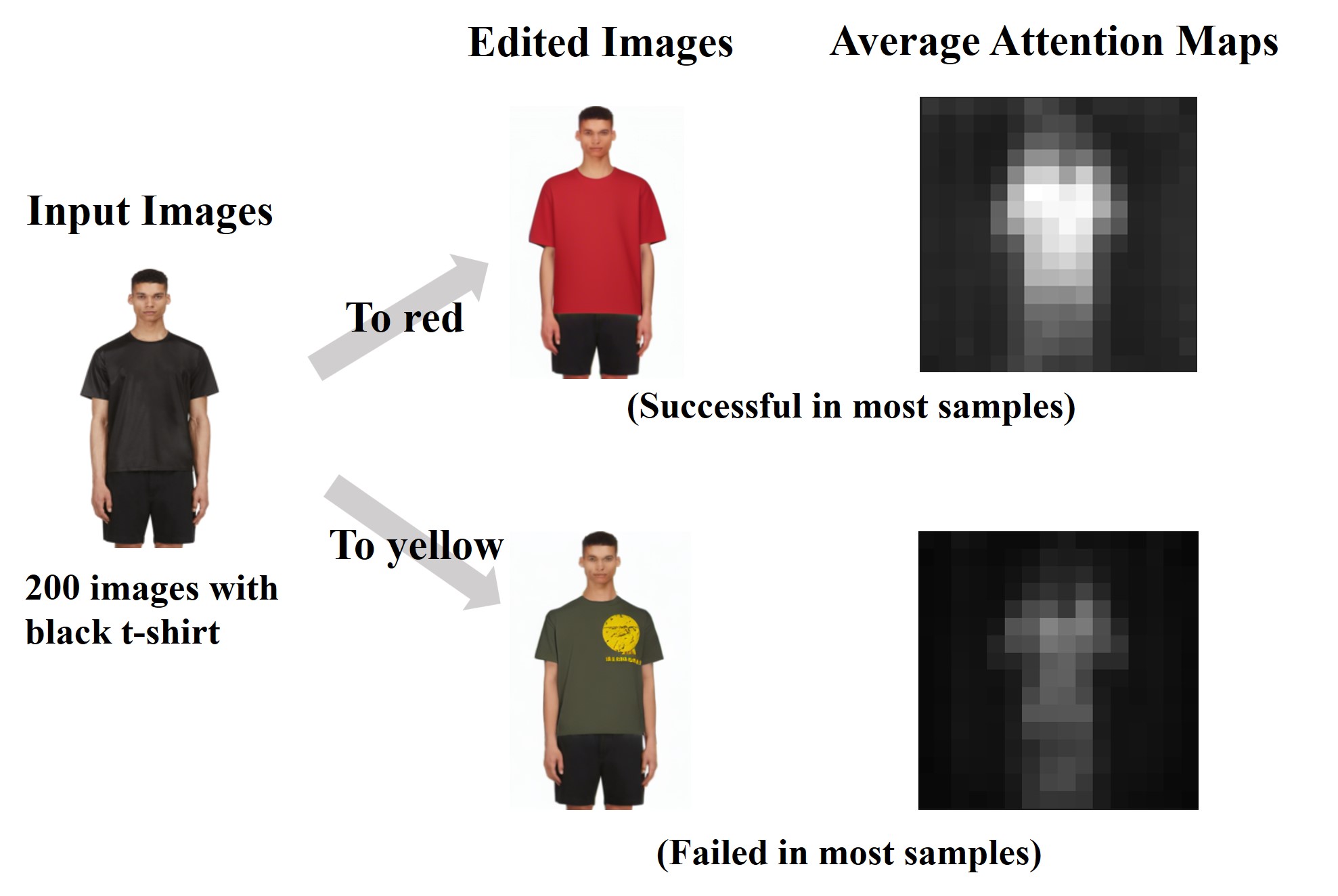}
    \vspace{-1.2em}
    \caption{Comparison of attention maps from successful and failed fashion editing cases. A total of 200 black t-shirts are edited to red and yellow, respectively, and the corresponding average attention maps are calculated.}
    \vspace{-1.5em}
    \label{fig:dis}
\end{figure}
\vspace{-0.25em}
\subsection{Image Editing with Attention-Enhanced Diffusion Model}
\vspace{-0.25em}
As shown in Fig. \ref{fig:intro}, when confronted with edits necessitating significant visual alterations, such as transitioning from a black t-shirt to yellow, many previous editing methods struggle to generate images that align with the target prompt. To identify the reasons of editing failures, an experiment, in which 200 black t-shirts are edited to red and yellow respectively, is conducted, and the corresponding average attention maps are calculated. As presented in Fig. \ref{fig:dis}, editing is more likely to succeed when the value of the attention map is high, and vice versa. Therefore, our method enhances editing magnitude by increasing the attention map values.



\noindent \textbf{Attention-Enhanced Diffusion Model.} Attention-Enhanced Diffusion Model is a training-free model,  which utilizes Stable Diffusion as its backbone. During editing process, the input image $x_{org}$ is first encoded into a latent space feature $x_0$ by the encoder. Then DDIM Inversion, starting from $x_0$, is executed:
\begin{equation}
{{x}_{t+1}}=\sqrt{1-{{\alpha }_{t+1}}}{{\epsilon }_{\theta }}({{x}_{t}},t)+\sqrt{{{\alpha }_{t+1}}}{{f}_{\theta }}({{x}_{t}},t),
  \label{eq1}
\end{equation}
where $t$ is the time step, ${\alpha }_{t}$ is a coefficient that decreases over time steps, ${{\epsilon }_{\theta}}$ is the noise prediction from the U-Net in diffusion model and ${{f}_{\theta }}$ is calculated by ${({{x}_{t}}-\sqrt{1-{{\alpha }_{t}}}{{\epsilon }_{\theta }})}/{\sqrt{{{\alpha }_{t}}}}$. The entire inversion process will take $S$ steps, encoding $x_0$ into ${x}_{S}$, which is a noise map containing spatial information of the input image. During this process, the DDIM trajectory ${{x}_{0}, {x}_{1}, ... ,  {x}_{S-1}}$ and the inversion noise map ${x}_{S}$ are collected.

Additionally, DDIM sampling, starting from a random noise ${x}_{T}^{no}$ with the same size as 
$x_0$, is also conducted:
\begin{equation}
{{x}_{t-1}^{no}}=\sqrt{\frac{{{\alpha }_{t-1}}}{{{\alpha }_{t}}}}\left( {{x}_{t}^{no}}-\frac{{{\alpha }_{t-1}}-{{\alpha }_{t}}}{{{\alpha }_{t-1}}\sqrt{1-{{\alpha }_{t}}}}{{\epsilon }_{\theta }}({{x}_{t}^{no}},t,c) \right),
  \label{eq2}
\end{equation}
where ${\alpha }_{t}$ is the same as in DDIM Inversion, $c$ represents the target prompt, and ${{\epsilon }_{\theta }}$ is the noise prediction from the U-Net. After ($T-S$) steps of DDIM sampling, ${x}_{T}^{no}$ will be denoised to ${x}_{S}^{no}$ and the attention map $A$ can be calculated by averaging all attention maps with resolution of 16$\times$16 in the U-Net.

After DDIM Inversion and sampling, the attention map $A$, noise maps ${x}_{S}$, ${x}_{S}^{no}$ and the mask $m$ from MaskNet are all input into the attention processor, in which a refined noise map ${x}_{S}^{re}$ with higher value of attention map can be obtained. 

\begin{figure}
    \centering
    \includegraphics[width=0.85\linewidth]{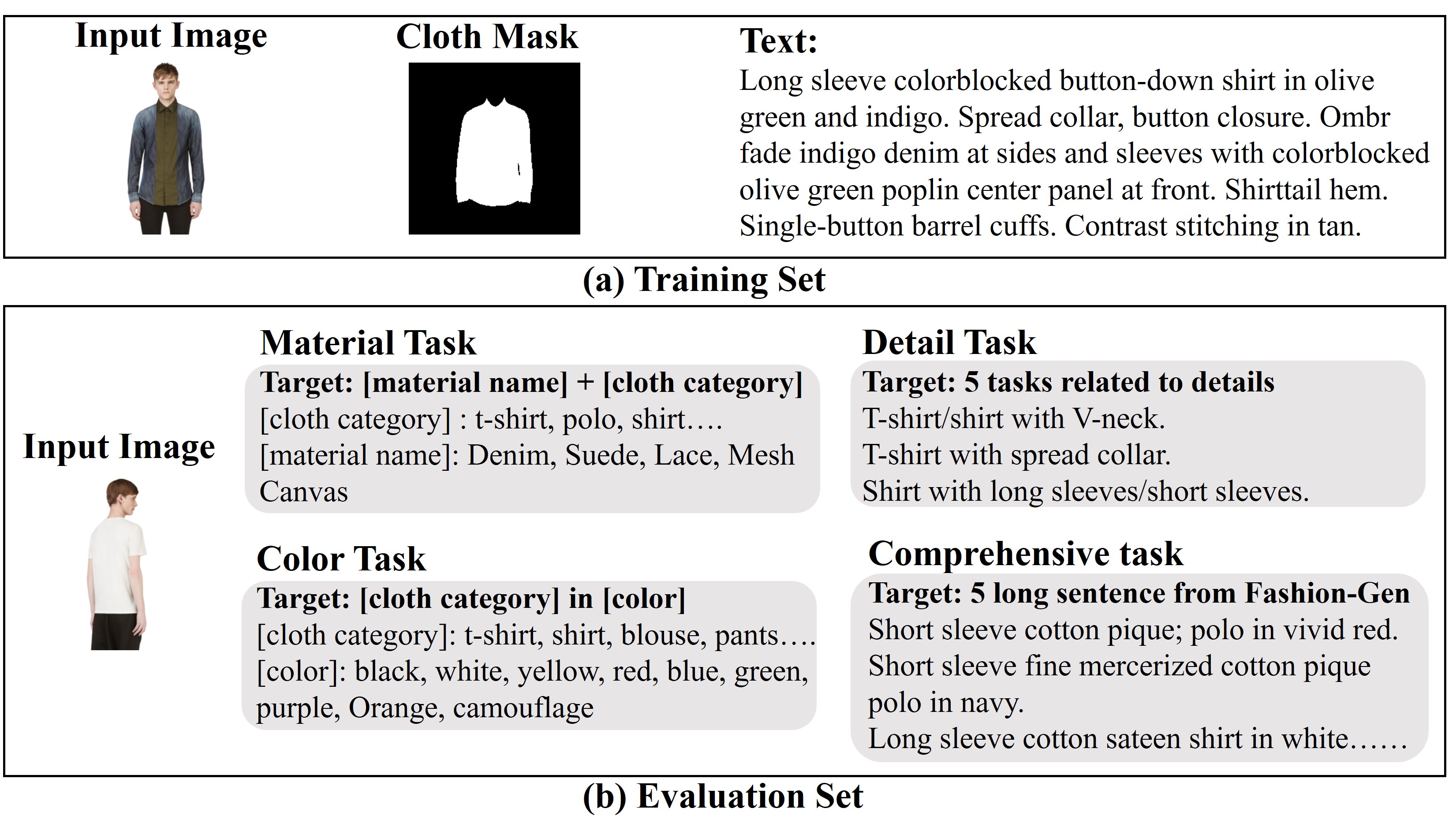}
    \vspace{-1.2em}
    \caption{Fashion-E dataset. (a) Training Set. The aligned image-text pair is used as the input, while the cloth mask is used as the ground truth for training MaskNet. (b) Evaluation Set. The input image and target text of the four fashion editing tasks are presented, which are used to evaluate editing models.}
    \vspace{-1.5em}
    \label{fig:dataset}
\end{figure}

Finally, guided by the target prompt, the refined noise map $x_{S}^{re}$ is utilized to conduct DDIM sampling, in which the intermediate results are blended with the DDIM trajectory:   
\begin{equation}
x_{t-1}^{re}=DDIM\left( x_{t}^{re} \right)\odot m+{{x}_{t-1}}\odot (1-m),
  \label{eq3}
\end{equation}
where $x_{t}^{re}$ is the refined noise map at time step $t$, $m$ is the mask from MaskNet. Applying Equation (\ref{eq3}) for $S$ steps, the refined noise map $x_{S}^{re}$ can be denoised to $x_{0}^{re}$. The edited image ${x_{out}}$ can be obtained by inputting $x_{0}^{re}$ into the decoder.

\begin{figure*}
    \centering
    \includegraphics[width=0.7\linewidth]{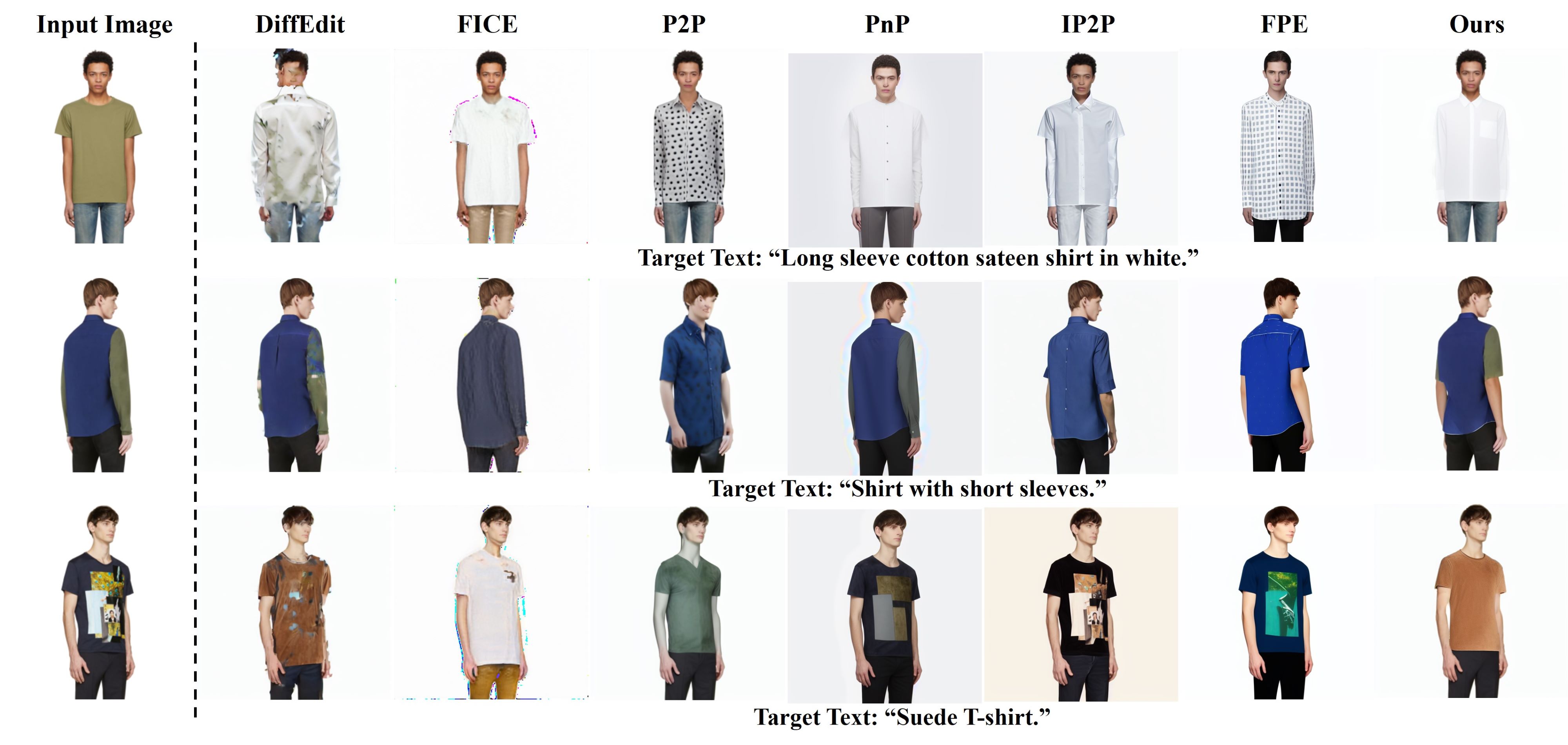}
    \vspace{-1.2em}
    \caption{Qualitative comparison with State-of-the-Art methods on Fashion-E. Target texts are placed below the row of images.}
    \vspace{-1.5em}
    \label{fig:result}
\end{figure*}

\noindent\textbf{Attention Processor.} The steps of Attention Processor are illustrated in Fig. \ref{fig:method}(c). Firstly, given the mask $m$ comprising $N$ pixels, the pixels inside the mask $m$ of the map $x_S$ can be represented as:
\begin{equation}
  {{G}_{ed}}=\{pix|{m}(pix)>0, pix\in {{x}_{S}}\},
\label{eq4}
\end{equation}
where $pix$ represents the pixel of noise map and ${G}_{ed}$ is a set of pixels within the mask $m$ in the noise map $x_S$. The Attention Processor then identify $N/2$ pixels with the largest values of attention map in ${{x}_{S}^{no}}$:
\begin{equation}
  {{G}_{pr}}=\{pix|{A}(pix)>{{V}_{min}},pix\in {{{x}_{S}^{no}}}\}, \\ 
\label{eq5}
\end{equation}
where ${G}_{pr}$ is a set of pixels with higher attention values from ${{x}_{S}^{no}}$ and ${V}_{min}$ is the ($0.5N$)th highest value in attention map $A$. Finally, all the pixels in ${G}_{ed}$ are substituted with the pixels in ${G}_{pr}$ in order. If all the pixels in ${G}_{pr}$ are used once, the surplus pixels in \({G}_{ed}\) are replaced by randomly selected ones from ${G}_{pr}$. 


\section{Experiments}
\vspace{-0.25em}
\subsection{Dataset and Metrics}
\vspace{-0.25em}
The Fashion-E dataset is constructed to comprehensively train and evaluate editing models. As shown in Fig. \ref{fig:dataset}(a), the training set of Fashion-E contains 29380 aligned image-text pairs sourced from the Fashion-Gen dataset, along with the corresponding cloth masks. With this set, the image-text pairs serve as input and the cloth masks serve as the ground truth for the training of MaskNet. Additionally, considering that existing fashion-related datasets cannot comprehensively evaluate models on different fashion tasks, we construct the evaluation set of Fashion-E. Comprising 2639 fashion images, the evaluation set is designed for four types of fashion editing tasks: color editing, detail editing, material editing, and comprehensive editing. Each sample in this set is annotated with a target text for a  specific editing task, as shown in Fig. \ref{fig:dataset}(b).

For the evaluation metrics, we employ CLIP Text Score (CLIP-T) \cite{ruiz2023dreambooth} to assess the alignment between edited image and target text, CLIP Image Score (CLIP-I) \cite{ruiz2023dreambooth} to evaluate the preservation of original image information, and LPIPS \cite{zhang2018unreasonable} to evaluate the perceptual similarity between the original and edited images.
\vspace{-0.25em}
\subsection{Comparison with State-of-the-Art Methods}
\vspace{-0.25em}
The proposed method is compared with six text-guided editing models on Fashion-E dataset: Diffedit \cite{couairon2023diffedit}, FICE \cite{pernuvs2023fice}, P2P \cite{hertz2023prompttoprompt}, PnP \cite{tumanyan2023plug}, IP2P \cite{brooks2023instructpix2pix} and FPE \cite{liu2024towards}. The experiments for these models are conducted based on their official implements.

\begin{table}[!htbp]
\centering
\vspace{-1em}
\caption{Quantitative comparison with State-of-the-Art methods on Fashion-E. Bold font indicates the best value for the metric, while underlined font indicates the second-best value.}
\resizebox{1\linewidth}{!}{
\begin{tabular}{cccccc}
\toprule
 Methods & Pub./Year & CLIP-T ($\uparrow$) & CLIP-I ($\uparrow$) & LPIPS ($\downarrow$) & Time ($\downarrow$)\\
\midrule
DiffEdit\cite{couairon2023diffedit}& ICLR$_{23}$ & 26.85 & 83.60 & 0.152 & 42.69\\
FICE\cite{pernuvs2023fice} & Arxiv$_{23}$ & 27.85 & 81.54 & 0.448 & 20.15\\
P2P\cite{hertz2023prompttoprompt} & ICLR$_{23}$ & 27.13 & 85.89 & 0.146 & 52.61\\
PnP\cite{tumanyan2023plug} & CVPR$_{23}$ & 28.02 & 87.77 & \underline{0.139} & 174.08\\
IP2P\cite{brooks2023instructpix2pix} & CVPR$_{23}$ & \underline{28.74} & \underline{87.85} & 0.167 & \textbf{6.01}\\
FPE\cite{liu2024towards} & CVPR$_{24}$ & 27.51 & 84.60 & 0.160 & 17.83\\
Ours& - & \textbf{29.20} & \textbf{90.37} &\textbf{0.137} & \underline{12.82}\\
\bottomrule
\end{tabular}
}
\label{table:result}
\end{table}

\noindent\textbf{Qualitative Comparisons.} Visualization results of different models are shown in Fig. \ref{fig:result}. 
As seen in the first and second rows of Fig. \ref{fig:result}, our method can accurately predict the editing regions, preserving more original information. Additionally, as shown in the first and third rows of Fig. \ref{fig:result}, our method demonstrates sufficient editing magnitude compared to other models, ensuring higher alignment between the edited image and the target prompt.

\noindent\textbf{Quantitative Comparisons.}  As presented in TABLE \ref{table:result}, our method achieves the highest CLIP-T score, indicating that our model has strongest editing capability. Our model also achieves significantly higher CLIP-I score and lowest average LPIPS score, which means our model can predict more accurate editing regions and achieve better results in local editing tasks. 
Our model is the second fastest among all models. While our model is slightly slower than IP2P, it significantly outperforms IP2P in terms of overall performance.
\begin{table}[h]
\centering
\vspace{-1em}
\caption{Ablation studies on Attention Processor (AP) and MaskNet.}
\resizebox{0.75\linewidth}{!}{
\begin{tabular}{cccccc}
\toprule
 MaskNet & AP & CLIP-T ($\uparrow$) & CLIP-I ($\uparrow$) & LPIPS ($\downarrow$)\\
\midrule
 & & 27.21 & 84.58 & 0.161\\
$\checkmark$ & & 27.11 & 89.76 & 0.133\\
 & $\checkmark$ & 28.74 & 89.03 & 0.151\\
$\checkmark$ & $\checkmark$ & 29.20 & 90.37 & 0.137\\
\bottomrule
\end{tabular}
}   
\label{table3}
\end{table}
\vspace{-1em}

\subsection{Ablation Study}
\par\noindent\textbf{MaskNet.} Ablating the MaskNet and using masks from local blending \cite{hertz2023prompttoprompt} in the diffusion Model lead to a notable decline in CLIP-I and LPIPS, as shown in the third line of TABLE \ref{table3}. This suggests that MaskNet improves the accuracy of editing regions and preserves original information outside the region.

\noindent\textbf{Attention Processor (AP).} Ablating the Attention Processor and directly using the inversion noise map during editing result in a significant drop in the CLIP-T score on Fashion-E, as seen in the second line of TABLE \ref{table3}. This indicates that the proposed Attention Processor significantly enhances the editing magnitude, aligning the edited image with the target prompt effectively.

\section{Conclusion}
This paper proposes a novel text-guided fashion image editing mothod named MADiff, which comprises two components: MaskNet and Attention-Enhanced Diffusion Model. The MaskNet addresses the issues of inaccurate localization of editing regions and Attention-Enhanced Diffusion Model enhances the editing magnitude. Quantitative and qualitative results on Fashion-E dataset indicate that our model outperforms other text-guided image editing methods in both text alignment and preservation of original information.

\bibliographystyle{IEEEbib}
\bibliography{refs}
\end{document}